\documentclass{article}

\usepackage{arxiv}

\usepackage[utf8]{inputenc} 
\usepackage[T1]{fontenc}    
\usepackage{hyperref}       
\usepackage{url}            
\usepackage{booktabs}       
\usepackage{amsfonts}       
\usepackage{nicefrac}       
\usepackage{microtype}      
\usepackage{lipsum}

\usepackage{graphicx}
\usepackage{stfloats}
\usepackage{multicol}

\usepackage{caption}
\usepackage{subcaption}
\usepackage{footnote}

\usepackage{listings}
\lstdefinestyle{mystyle}{
    breaklines=true,                 
    captionpos=t,
}

\usepackage[stable]{footmisc}

\title{Neural Academic Paper Generation}

\author{
  Samet Demir\\
  Department of Computer Engineering\\
  Boğaziçi University\\
  İstanbul, Turkey \\
  \texttt{samet.demir1@boun.edu.tr} \\
   \And
  Uras Mutlu\\
  Department of Computer Engineering\\
  Boğaziçi University\\
  İstanbul, Turkey \\
  \texttt{uras.mutlu@boun.edu.tr} \\
    \And
  Özgür Özdemir\\
  Department of Computer Engineering\\
  Istanbul Bilgi University\\
  İstanbul, Turkey \\
  \texttt{ozgur.ozdemir@bilgiedu.net} \\
}

\date{}

\begin{document}
\maketitle

\begin{abstract}
In this work, we tackle the problem of structured text generation, specifically academic paper generation in \LaTeX{}, inspired by the surprisingly good results of basic character-level language models. Our motivation is using more recent and advanced methods of language modeling on a more complex dataset of \LaTeX{} source files to generate realistic academic papers. Our first contribution is preparing a dataset with \LaTeX{} source files on recent open-source computer vision papers. Our second contribution is experimenting with recent methods of language modeling and text generation such as Transformer and Transformer-XL to generate consistent \LaTeX{} code. We report cross-entropy and bits-per-character (BPC) results of the trained models, and we also discuss interesting points on some examples of the generated \LaTeX{} code. 
\end{abstract}

\section{Introduction}
Even with the recent advances on natural language generation (NLG) methods that employ deep learning architectures, it is still a challenge to generate semantically consistent and structured essays or texts in general. Therefore, academic paper generation is still a compelling problem on vast NLG field.

Almost all academic papers in the field of computer science, including this one, are written in \LaTeX{} typesetting system \cite{1986lamport}. \LaTeX{} has many syntactic rules to create objects such as headers, figures, and tables. Therefore, in an academic paper written in \LaTeX{} format, there are both semantic and syntactic long term dependencies inside the text. The semantic dependency, which refers for consistency on the flow of the text, is aimed to maintain while writing the papers. For instance, the introduction and conclusion are consistent with the subject and the motivation of the work, as well as the rest of the paper. In addition to the semantic integrity of the essay, the syntactic dependency on keywords of \LaTeX{} is needed to be sustained to compile the file successfully. There can be a complex table or a section with many subfigures, and sometimes it is not easy to keep up with the brackets and special keywords needed to create a table successfully. Both of these long term dependencies are a challenge for an automated academic paper generation system.

Fortunately, there are many recent studies on natural language generation focusing on generating realistic texts by taking long term dependencies into account \cite{Attention_is_All_you_Need,devlin2018bert,transformer_xl,radford2019language,ijcai2018-567}. There are also widely used machine learning methods such as Recurrent Neural Networks (RNN), and Long-Short Term Memory (LSTM) \cite{Hochreiter:1997:LSM:1246443.1246450} that are designed to handle short and long term dependencies in accordance with any data involving such dependencies. Our primary motivation is to use some of the recent methods with a new dataset consisting of academic papers written in \LaTeX{} to see if the successful NLG models can also accomplish to generate realistic academic papers.

Rest of the paper is organized as follows: The literature on academic paper generation and character-level text generation in general is given in Section \ref{sec:prev}; details of the dataset and the models are in Section \ref{sec:method}; Section \ref{sec:results} contains the experimental setup and the results, and Section \ref{sec:conc} contains the conclusion and future work.

\section{Previous Work}
\label{sec:prev}

To the best of our knowledge, there are only a few works on automated academic paper generation. One of such works is SCIgen \cite{SCIgenAnAutomaticCSPaperGenerator} that generates random sentences, graphs, and citations from a handwritten context-free grammar. \cite{SCIgenAnAutomaticCSPaperGenerator} stated that some generated papers by the tool had been accepted to a few conferences. Our work aims to use modern machine learning and language modeling techniques, rather than using a handwritten grammar, to generate more realistic academic papers.

Another work by \cite{char_rnn} showed that even simple RNN models capture interesting features and have the ability to generate any text by learning character-level language models. \cite{char_rnn} used the \LaTeX{} source files of a linear algebra book to generate mathematical formulas and proofs. It is claimed that it was possible to compile the generated \LaTeX{} code with little post-processing. In this work, we take a few steps further by training some recent models with a relatively large amount of \LaTeX{} data in order to be used in generating fake academic papers.

On general language modeling and text generation literature, many recent studies improve state-of-the-art. In one of these studies, \cite{Attention_is_All_you_Need} introduced Transformer which consists of a mechanism called \textit{attention} and the model aims to focus on specific parts of the sequential data for language modeling and language understanding. Since it is hard to train sequential models in a parallel manner, attention mechanism completely eliminates the recurrence of the model and paves the way to much faster training with parallel GPUs while improving the success of earlier models. In Transformer architecture, the sequence information is provided with positional encodings, and the model is expected to learn the joint probability of given sequences by using an encoder-decoder network.

Another model, namely Transformer-XL \cite{transformer_xl}, is proposed to eliminate the drawback due to the usage of the fixed-length context vector in Transformer. In other words, Transformer architecture uses a fixed size context vector, yet Transformer-XL extends this idea to employ a variable-length context vector to learn the dependencies in the data by also preserving the temporal structure. It is shown that Transformer-XL learns longer dependencies than both RNNs and Transformers.

There is also the GPT-2 model \cite{radford2019language} that recently set the state-of-the-art on language modeling by training a Transformer with $1.5$B parameters and a huge dataset called WebText. The outstanding results showed that it is possible to drastically improve the quality of language models by using more data and more parameters.

\cite{ijcai2018-567} also studied on language modeling, yet a more challenging task, such as writing an essay paragraph consisting of selected topics. They used an augmented LSTM structure for generating the text sequences that each of them are related to the topics as well as before sequences. Although they showed how difficult to sustain the integrity of the writing, they managed to outperform other essay generation studies on both quantitative and qualitative results.  

\section{Methodology}
\label{sec:method}

In this work, we aim to generate academic papers in \LaTeX{} since it is a widely used typesetting system in the academia and it is self-contained such that it can be used to write academic papers by generating plain text. The process of preparing the new dataset composed of papers written in \LaTeX{} is described in Section \ref{sec:dataset}. Because determining word boundaries for \LaTeX{} format is difficult, we approached the problem as a character-level language modeling problem. Besides, the diversity of vocabulary used in academic papers is extensive. Therefore the computational complexity would have been higher if word-level language modelling was used rather than character-based.

In experiments, the performance of Transformer and Transformer-XL are compared adversarial to the baseline RNN architecture which exploits LSTM structure. As commonly used in language modelling  \cite{graves2013generating,transformer_xl},
the models that are explained in Section \ref{sec:models} are used to predict $P(X_t|X_1,X_2,...,X_{t-1})$ and overall learn the joint probability $P(X_1,X_2,...,X_T)$ where $X$ is the given sequence of characters.

\begin{table}[hbt!]
\centering
\begin{tabular}{|c|c|}
\hline
\textit{\textbf{Feature}} & \textit{\textbf{Value}} \\ \hline
unique tokens & 102 \\ \hline
total tokens & 37,921,928 \\ \hline
papers & 799 \\ \hline
\end{tabular}
\caption{Some quantitative information about the dataset.}
\label{tab:dataset}
\end{table}

\subsection{Dataset\footnote{We decided not to share the dataset because of ethical concerns. However, our code, which is shared, can be used to recreate similar datasets.}}

\label{sec:dataset}The dataset is prepared from academic papers on arXiv \cite{arxiv} since there did not exist a similar dataset which consists of academic papers written in \LaTeX{} to the best of our knowledge. Table~\ref{tab:dataset} shows some quantitative information about the dataset.
\subsubsection{Preparation}
Following steps are completed in order to create the dataset consisting of \LaTeX{} files;
\begin{enumerate}
    \item Academic papers on arXiv which are tagged as Computer Vision and submitted between 2015 - 2018 are selected as a subset of academic papers to base the dataset on.
    \item The source files of the selected academic papers are downloaded.
    \item Each paper which initially consists of multiple files is compiled into one \LaTeX{} file.
\end{enumerate}

\subsubsection{Preprocessing}
The raw dataset is preprocessed on character-level in order to remove the noise coming from \LaTeX{} comments. The infrequent characters that appeared less than 100 times in the whole dataset are deleted. The cleaned dataset which consists of multiple \LaTeX{} files is concatenated into one file to feed the models. The resulting sequence is segmented into same length sequences, and these sequences form the batches of the same size.

\subsection{Models}
\label{sec:models}
In this study, a character based LSTM structure is selected as the baseline model since \cite{char_rnn} showed its performance in \LaTeX{} generation. We also experimented with the Transformer \cite{Attention_is_All_you_Need} because of its recent achievements on sequence modeling. Finally, We experimented with Transformer-XL \cite{transformer_xl} since it is claimed that the model is able to capture long term dependencies better than the others. Char-LSTM, Transformer, and Transformer-XL models are described in Subsections \ref{sec:rnn}, \ref{sec:transformer}, and \ref{sec:transformer-xl} respectively.

\begin{figure}[hbt!]
    \centering
    \includegraphics[width=0.4\textwidth]{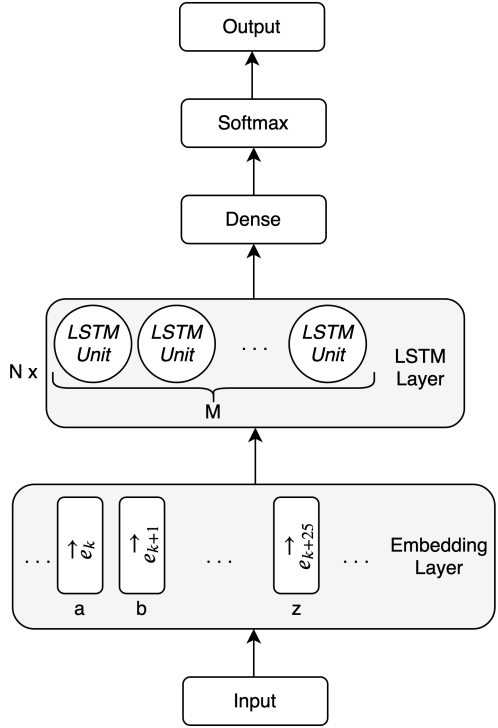}
    \caption{Char-LSTM model that is used in the experiments, where $\vec{e_k}$ is the embedding vector for the $k$-th character, N is the number of LSTM layers and M is the number of recurrent units.}
    \label{fig:rnn}
\end{figure}

\subsubsection{RNN: Char-LSTM}
\label{sec:rnn}

RNN architectures provide intelligent networks to learn sequential information of inputs, and RNNs became a popular approach in sequence modelling \cite{Sutskever2014,char_rnn,Sutskever:2011:GTR:3104482.3104610,graves2013generating}. However, \cite{Hochreiter:91} and \cite{Bengio:1994:LLD:2325857.2328340} described the vanishing gradients problem that makes the RNN training difficult. Thus, new RNN architectures such as LSTM \cite{Hochreiter:1997:LSM:1246443.1246450} and GRU \cite{cho-etal-2014-properties} have been developed to avoid vanishing gradients issue and make RNNs practically more useful. 
\cite{char_rnn} trained a multi-layer LSTM structure on \LaTeX{} source file of an Algebraic Geometry book and evaluated the \LaTeX{} generation performance of the model, so it inspired us to perform this study, as well as providing a baseline model. In this baseline model, which we will call Char-LSTM for the rest of this study, a character sequence input is processed by an embedding layer initially. The layer maps the characters to a fixed-size embedding vector. Embedding vectors are then fed to sequentially connected LSTM layers. LSTM layer in step $t$ while processing the given sequence could be mathematically described as $h_t = LSTM(h_{t-1}, x_t)$ where $h_t$ is the hidden state and $x_t$ is the input at the current step. The output of the LSTM layers then given to a dense layer. Lastly, softmax function is applied to the output vector $x$ of the dense layer. A flow chart of the model is given in Figure \ref{fig:rnn}.

\subsubsection{Transformer}
\label{sec:transformer}

\begin{figure*}[hbt!]
    \centering
    \includegraphics[width=0.95\textwidth]{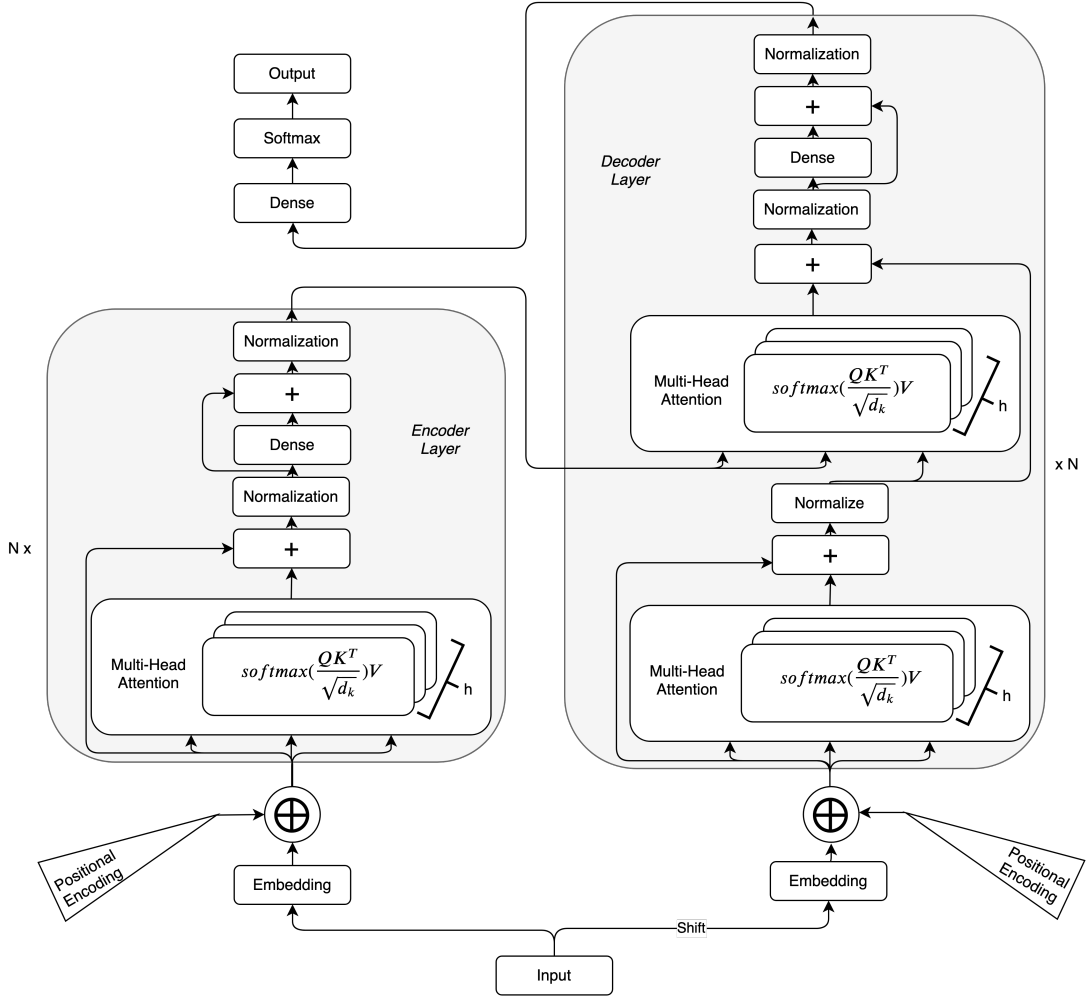}
    \caption{Transformer model, where $h$ is the number of self attention heads in a multi-head attention layer, $N$ is the number of hidden encoder and decoder layers, $\oplus$ is the concatenation layer, + is the vector addition layer, arrows($\uparrow$) refer to the flow of data, $Q$,$K$ and $V$ are abbreviations for $X^QW^Q_i$, $X^KW^K_i$ and $X^VW^V_i$ respectively, $X^Q$, $X^K$ and $X^V$ are the corresponding inputs and $i$ is the index of a self attention head in a multi-head attention layer.}
    \label{tranformer-fig}
\end{figure*}

Recently, attention mechanisms are integrated into RNNs in order to allow modeling distance-free dependencies \cite{bahdanau2014neural,Kim2017StructuredAN}. Transformer \cite{Attention_is_All_you_Need} has been introduced as a sequence transduction model which is based on entirely attention mechanisms without recurrence such that the Transformer could make more use of parallel processing than the recurrent networks. In recent studies \cite{devlin2018bert,radford2019language}, Transformer and its components indicated significant success and became a popular approach on sequence modeling. Therefore, it is a prominent candidate model for academic paper generation problem. 

The Transformer consists of embedding, positional encoding, multi-head attention, and other standard building blocks of deep learning architectures, such as normalization \cite{Ioffe:2015:BNA:3045118.3045167} and dense layers. Positional encoding is used in order not to lose the positional information of the given sequence. Multi-head attention is the layer which contains multiple self attention components connected in parallel. Self attention is described mathematically as
$$softmax(\frac{(X^QW^Q)(X^KW^K)^T}{\sqrt{d_k}}) (X^VW^V)$$ 
where $X = [X^Q, X^K, X^V]$ is the input to the self-attention, $W^Q$, $W^K$ and $W^V$ are parameter matrices used to project input to queries, keys, and values respectively and $d_k$ is the number of dimensions of the keys. The input feeds through an embedding layer, a concatenation layer where positional encoding is concatenated, and encoder layers. Furthermore, shifted input goes through a similar embedding layer, and decoder layers.

An encoder layer consists of a multi-head attention layer, an addition layer where input and output of the multi-head attention are summed just like a residual connection in ResNet \cite{DBLP:journals/corr/HeZRS15}, a normalization, a dense layer, an addition layer where input and output of the dense layer are summed, and finally a normalization layer in the given order.

A decoder layer consist of a multi-head attention layer, an addition layer where input and output of multi-head attention is summed, a normalization, a multi-head attention layer again where at this time input for query ($X^Q$) and key ($X^K$) come from the output of encoder layers and value ($X^V$) comes from the output of the normalization, dense, and addition layers where input and output of the dense layer is summed, and finally a normalization layer in the given order. At the end, the output of the decoder layers pass through a dense layer and a softmax layer. Figure \ref{tranformer-fig} shows the complete model and \cite{Attention_is_All_you_Need} describes the details of the model.

\subsubsection{Transformer-XL}
\label{sec:transformer-xl}
The Transformer model makes use of the constant number of context tokens since the model takes fixed-size sequences. The context length is generally selected as few hundreds in practice because of the computational limitations. In addition to these limitations, Transformer architecture also has no ability to carry information between context segments of the sequences. The problem that generally occurs in practice is to segment a given sequence to vectors consisting of fixed-sized context tokens without respecting semantic boundaries, which is also a problem about long term dependencies. Similarly, \LaTeX{} includes long term dependencies such as the dependency between \texttt{\textbackslash begin} and \texttt{\textbackslash end} statements since there could be a long text in-between.

Besides, the study done by \cite{khandelwal-etal-2018-sharp} showed that LSTM language models have the capacity of using 200 context tokens on average. Intuitively, a model without ability to learn interconnections on segments of the sequences would not be sufficient to generate \LaTeX{} files that desires longer dependencies successfully. Therefore, a model that handles longer dependencies between sequences become more appropriate solution against Char-LSTM and Transformer on \LaTeX{} generation task.

\begin{table}[!hbt]
    \centering
    \begin{tabular}{|c|}
    \hline
    {\small Dependencies in Transformer} \\
    \includegraphics[width=0.45\textwidth]{transformer-xl.png}
    \\ \hline \hline
    {\small Dependencies in Transformer-XL} \\
    \includegraphics[width=0.45\textwidth]{transformer-xl2.png}
    \\ \hline
    \end{tabular}
    \caption{Example two-layer illustration of the Transformer-XL in comparison to Transformer, where $x_t$ is the input at time $t$, $h_t^n$ is the hidden state on layer $n$ at time $t$, arrows($\uparrow$) represent dependencies, $x_{a:b}$ is abbreviation for $[x_a,x_{a+1},\dots,x_b]$, dashed line in Transformer figure represent no information flow in-between, and dashed arrows in Transformer-XL figure show the newly added dependencies.} 
    \label{tab:transformer-xl}
\end{table}

Recently, \cite{transformer_xl} addressed the mentioned dependency problems and introduced Transformer-XL, an extended version of Transformer which stores and makes use of previous hidden states so that it increases the capacity of the model to capture long term dependencies. The main extension made by \cite{transformer_xl} to Transformer is to change the self attention layers as follows:
$$\hat{h}_{t+1} = h_{t} \oplus h_{t+1}$$
$$softmax(\frac{(h_{t}W^Q)(\hat{h}_{t}W^K)^T}{\sqrt{d_k}}) (\hat{h}_{t}W^V)$$ 
where $h_{t}$ is the input for the current self attention layer and the hidden state of the previous layer for the $k$-th input segment $[X_{t*l+1},X_{t*l+2},...,X_{(t+1)*l}]$ of fixed-size $l$ , $\hat{h}_{t}$ is the extended version of $h_{t}$ and $\oplus$ is the concatenation operator. Transformer-XL also includes a new more suitable positional encoding since the positional encoding which is introduced in \cite{Attention_is_All_you_Need} fails to differentiate $X_{t,j}$ and $X_{t+1,j}$, where $X_{t,j}$ is the $j$-th token in $t$-th input segment. 

\section{Experiments\footnote{The code for the experiments and the dataset can be found at \href{https://github.com/inzva/fake-academic-paper-generation}{https://github.com/inzva/fake-academic-paper-generation}}}
\label{sec:results}

\begin{savenotes}
\begin{table*}[hbt!]
\centering
\begin{tabular}{|c||c||c||c|}
\hline
\textit{\textbf{Hyperparameter}} & \textit{\textbf{Char-LSTM}} & \textit{\textbf{Transformer}} & \textit{\textbf{Transformer-XL}} \\ \hline
sequence length & 100 & 128 & variable* \\ \hline
batch size & 64 & 4096 & 22 \\ \hline
hidden layers (N) & 1 & 2 & 12 \\ \hline
embedding size & 256 & 256 & 512 \\ \hline
hidden size & 1024 & 256 & 512 \\ \hline
number of heads (h) & - & 4 & 8 \\ \hline
dropout rate & - & 0.1 & 0.1 \\ \hline
optimizer & adam & adam & adam \\ \hline
learning rate schedule & - & custom** & cosine \\ \hline
learning rate & 0.001 & 0.2 & 0.00025 \\ \hline
beta1 (adam) & 0.9 & 0.9 & 0.9 \\ \hline
beta2 (adam) & 0.999 & 0.997 & 0.999 \\ \hline
epsilon (adam) & $e^{-8}$ & $e^{-9}$ & $e^{-8}$ \\ \hline
\end{tabular}
\caption[]{Hyperparameters used for the models. *The randomized sequence length option provided by \cite{transformer_xl} is selected. **The default scheduler parameters is used as follows ${l_{schedule}=linear\_warmup\cdot decay^{-0.5}\cdot linear\_decay}$.}
\label{tab:hyperparams}
\end{table*}
\end{savenotes}

\begin{figure*}[hbt!]
    \centering
    \begin{subfigure}{.3\textwidth}
        \centering
        \includegraphics[width=\textwidth]{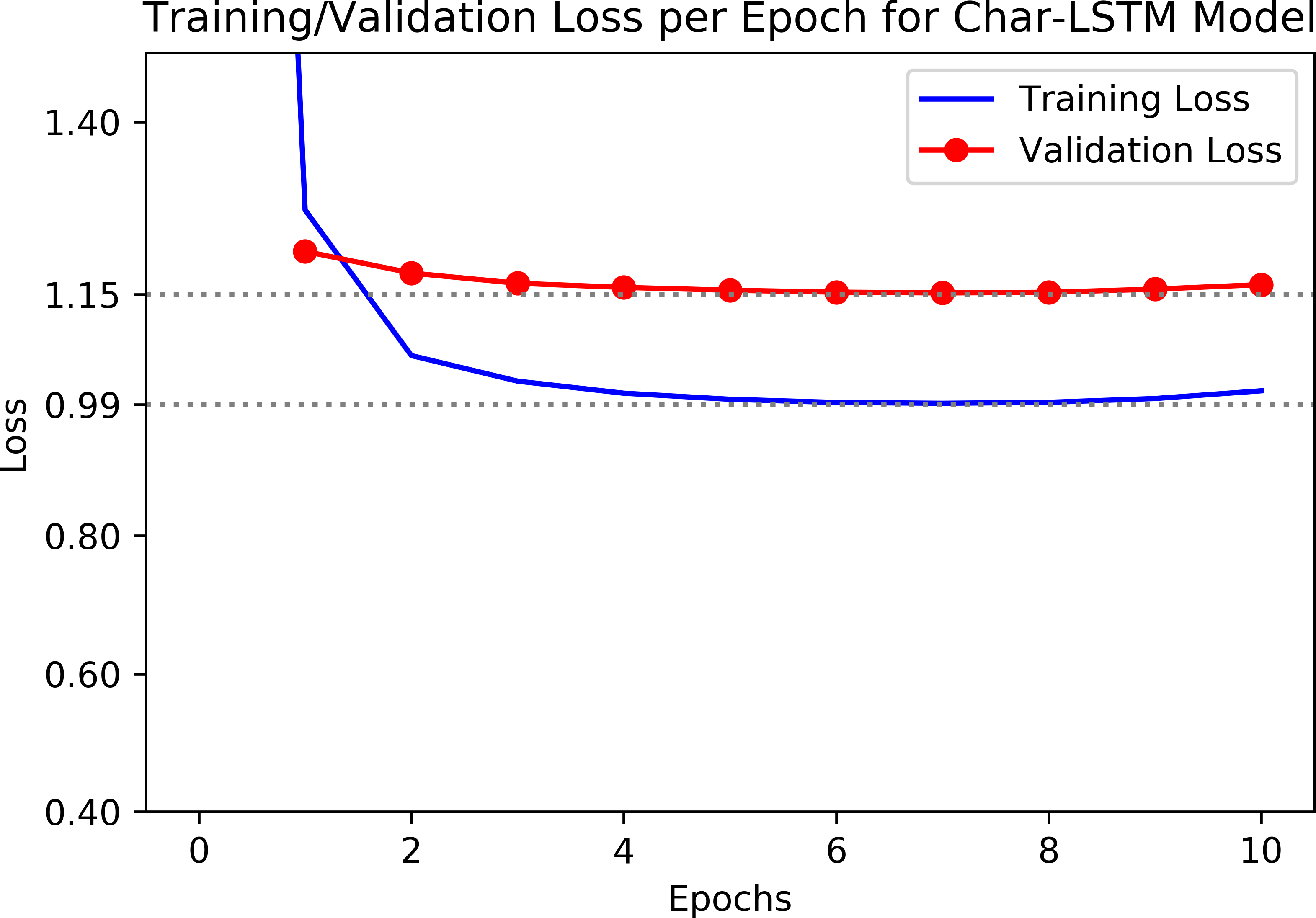}
        \caption{Char-LSTM}
        \label{fig:char-lstm-loss}
    \end{subfigure}
    \begin{subfigure}{.3\textwidth}
        \centering
        \includegraphics[width=\textwidth]{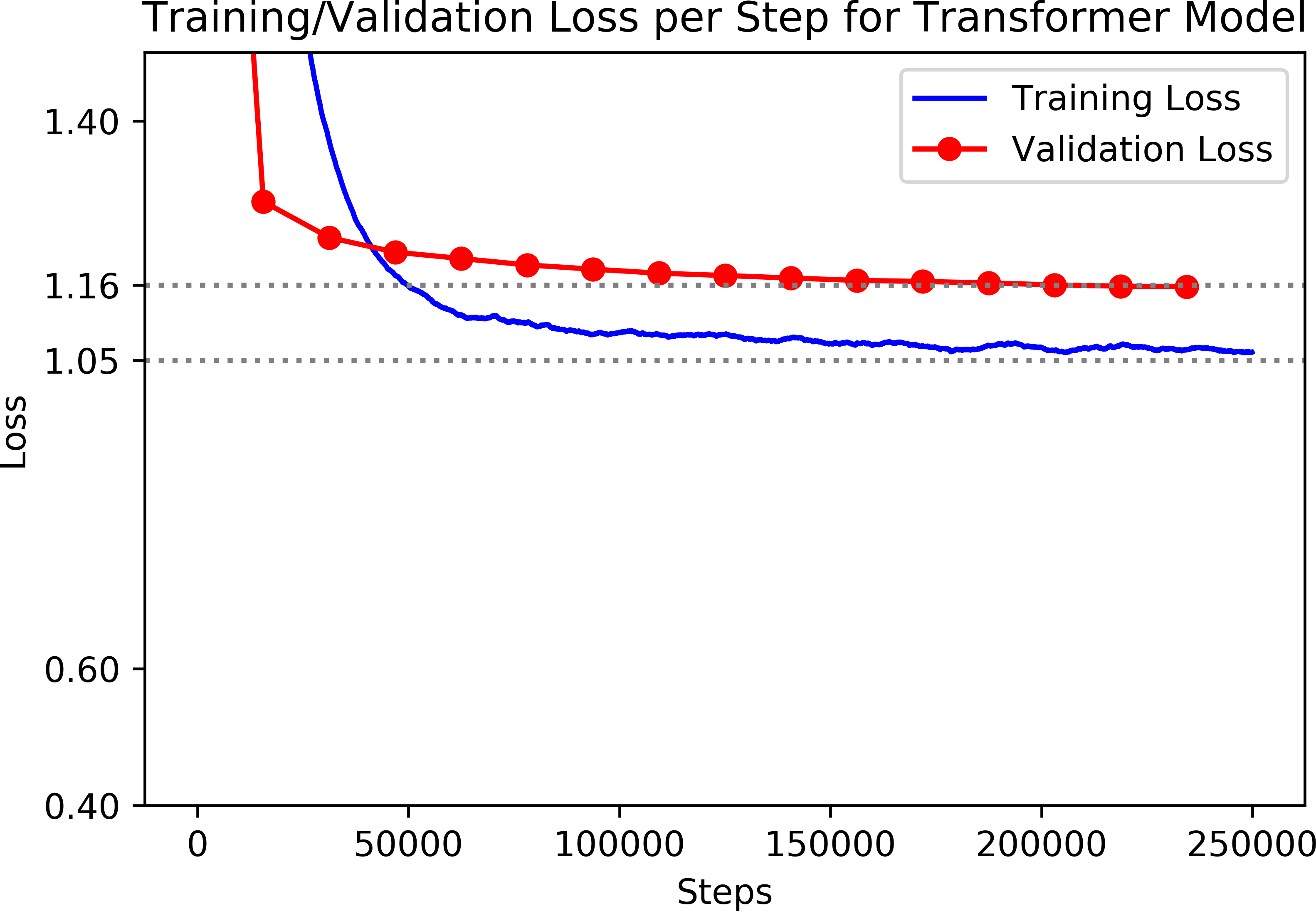}
        \caption{Transformer}
        \label{fig:transformer-loss}
    \end{subfigure}
    \begin{subfigure}{.3\textwidth}
        \centering
        \includegraphics[width=\textwidth]{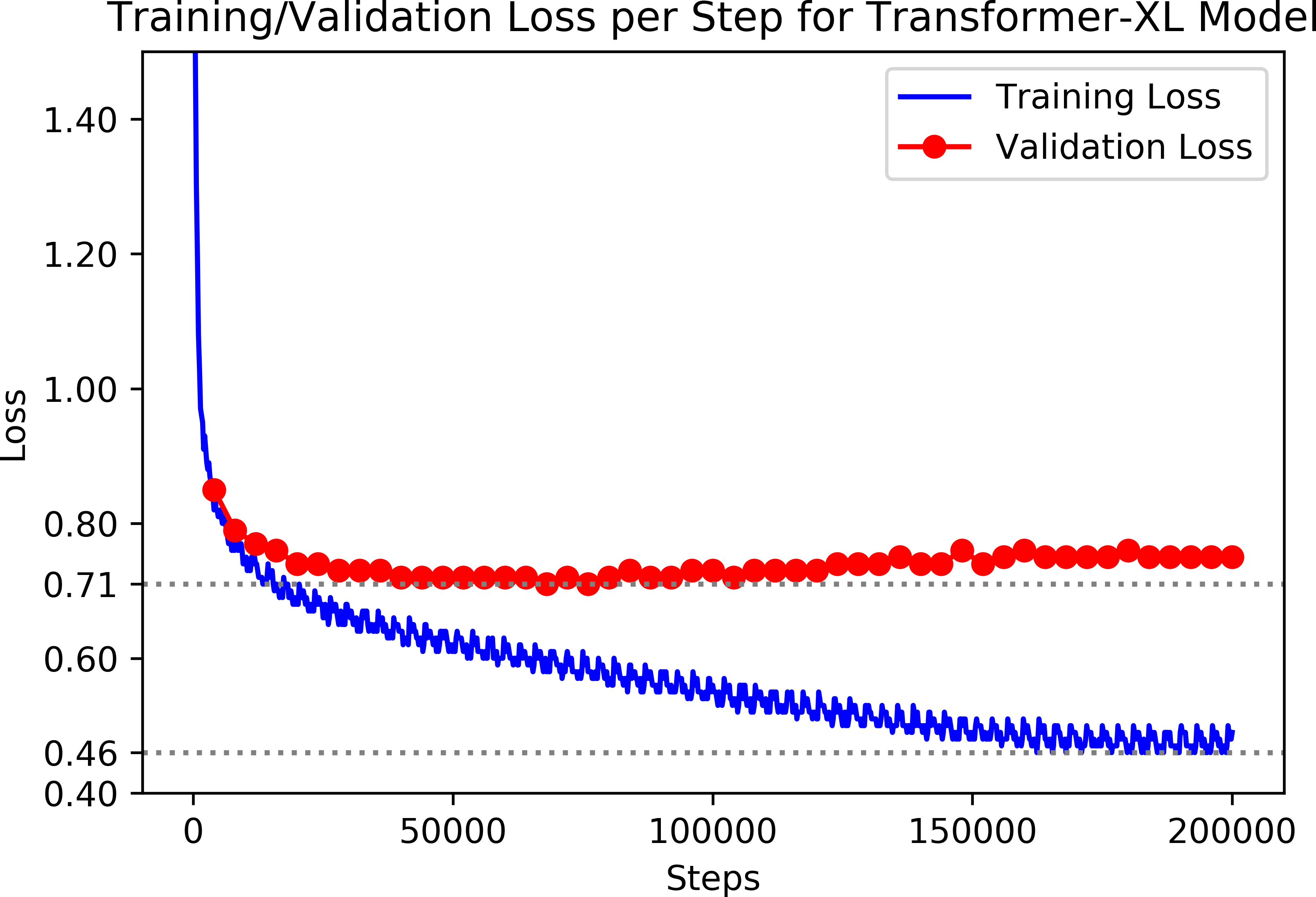}
        \caption{Transformer-XL}
        \label{fig:transformer-xl-loss}
    \end{subfigure}
    \caption{Training and validation loss per epoch/step for each models. Each epoch in Char-LSTM refers to  $\sim 20$K steps.}
    \label{fig:training_losses}
\end{figure*}

Implementations provided by the authors of the selected models are used in the experiments. TensorFlow \cite{tensorflow2015-whitepaper}, Keras \cite{chollet2015keras}, PyTorch \cite{paszke2017automatic} and Tensor2Tensor \cite{tensor2tensor} are used since the models are originally implemented using one or more of these. 

\subsection{Experimental Setup}
\label{sec:experimental_setup}
We used three different computational resources to train different models. Char-LSTM and Transformer models are trained on a Tesla K80 GPU, while the Transformer-XL model is trained using four Tesla V100 GPUs in parallel.

Hyperparameters used for Char-LSTM, Transformer, and Transformer-XL models can be seen in Table \ref{tab:hyperparams}. For Char-LSTM model, we experimented with different values for context vector length, hidden layer size, and hidden unit size. For the sake of simplicity, we decided to train the model with only one hidden layer as our baseline. For Transformer model, we used the default hyperparameter settings given in Tensor2Tensor \cite{tensor2tensor}. Hyperparameters for the Transformer-XL model is also chosen as the base hyperparameter setting version of the model in the original work by \cite{transformer_xl}. This model is given in the official code written by the authors of the paper. The base hyperparameter setting of Transformer-XL is more shallow and simpler than the actual hyperparameter setting and the reason why we choose this setting is the lack of computational resources since the actual hyperparameter setting makes the model more complex and it is originally trained on a TPU cluster.

For text generation on trained models, we experimented with different values of the softmax \textit{temperature} parameter. The temperature parameter is used for controlling the randomness of the outputs by scaling the values of the softmax outputs. In other words, computed logits of the softmax layer are divided by the temperature. As the temperature value approaches $1.0$, final output converges to the actual values of the logits and this makes the model more ``random'' during sampling since the probabilities are more evenly distributed. When the temperature goes to $0$, this favors the logits with higher values and the model is more ``confident'' but also ``conservative'', always choosing the very likely outcomes during sampling from the output probabilities.

\subsection{Results}

\begin{table*}[hbt!]
    \begin{center}
        \begin{tabular}{|c||c|c||c|c|}
            \cline{1-5}
     		\textit{\textbf{Model}}  &  \textit{\textbf{Training CE}} & \textit{\textbf{Validation CE}} & \textit{\textbf{Training BPC}} & \textit{\textbf{Validation BPC}}  \\ \hline\hline
            Char-LSTM & 0.99 & 1.15 & 1.43 & 1.66  \\ \hline
            Transformer & 1.05 & 1.16 & 1.51 & 1.67 \\ \hline
            Transformer-XL & \textbf{0.48} & \textbf{0.71} & \textbf{0.69} & \textbf{1.02} \\ \hline
        \end{tabular}
        \caption{Best training and validation cross entropy (CE) errors and bits-per-character (BPC) calculations on trained models.}
        \label{tab:cross_entropy}
    \end{center}
\end{table*}

\lstset{
    basicstyle=\footnotesize\rmfamily,
    style=mystyle,
    language=[LaTeX]{TeX},
    frame=topline,
    aboveskip=-1em,
    belowskip=-0.8em,
    morekeywords={\$},
}
\begin{table*}[!hbt]
    \centering
    \begin{tabular}{|p{0.47\linewidth}||p{0.47\linewidth}|}
    \hline
    {\footnotesize
\begin{lstlisting}[caption=Passage generated by Char-LSTM]
 
\end{lstlisting}
    We further analyze the effect of obtaining the proposed approach of our method and fine-tuning the spatial relationship between the surface instead of the annotation error \textbf{(as provided by the original image)} which are shown in Fig.~\textbf{\textbackslash ref\{fig:runtime\}}. The second step is trained on the training set in a standard environment with a single sample image of the same size with a single color \textbf{(blue)} are present in the image. We then evaluated the performance of the pre-trained CNN features to obtain better performance than the state-of-the-art face recognition model in the appendix. \label{lis:lstm-paragraph}
    }
    &
    {\footnotesize
\begin{lstlisting}[caption=Passage generated by Transformer-XL]
 
\end{lstlisting}
    We verify the effectiveness of our proposed method on both the 50k \textbf{\textbackslash emph\{and\}} the 2012 dataset available, all reported a reference performance of \textbf{\$66.1\%\$} on the validation set. We also drop our initial performance (denoted as RPN \textbf{\$\textbackslash stackanchor\{+\}\{\}\$}) and add RRCN \textbf{\$\textbackslash stackanchor\{+\}\{\}\$} to get an average performance according to the three evaluation metrics. Results are shown in Table~\textbf{\textbackslash ref\{tab:accuracy\}}. The three evaluation metrics are measured by computing the average, and showing the difference of RRCN with respect to the obtained initial performance (denoted as \textbf{\textbackslash textit\{initial performance\}}).}\newline
    \label{lis:transformer-xl-paragraph}
    \\ 
    \hline
    \end{tabular}
    \caption{Comparison of generated example passages, on which interesting points are highlighted.}
    \label{tab:output-comparison2}
\end{table*}

\lstset{
    basicstyle=\footnotesize,
    style=mystyle,
    language=[LaTeX]{TeX},
    frame=topline,
    aboveskip=-1em,
    belowskip=-1em,
    morekeywords={},
}
\begin{table*}[!hbt]
    \centering
    \begin{tabular}{|p{0.47\linewidth}||p{0.47\linewidth}|}
    \hline
\begin{lstlisting}[caption=\LaTeX{} figure generated by Char-LSTM]
\begin{figure*}[t]
\centering
\includegraphics[scale=0.9]{fine-tuning.pdf}
\captionsetup{def}{
\begin{figure}
\begin{subfigure}{.2\columnwidth}
\centering
\includegraphics[width=0.15\textwidth]{figure/4.jpg}}
\hyphil_im_impool/serre/budget & DBN \\
\hline
MNIST & Last y3\\
\hline
\end{tabular}}
\end{lstlisting}
    &
\begin{lstlisting}[caption=\LaTeX{} figure generated by Transformer-XL]
\begin{figure}[b!]
\centering
\includegraphics[width=0.99\linewidth]{Fig_fig_arch_fig_v2.pdf}
\caption{\small{Architecture of the VGGNet  (Siamese net), as illustrated in Figure~\ref{fig_arch}. The architectures have indeed the same architecture as VGGNet.} }
\label{fig_arch}
\end{figure*}
\end{lstlisting} 
    \\ 
    \hline
    \end{tabular}
    \caption{Comparison of generated example \LaTeX{} figures, on which basic \LaTeX{} keywords are highlighted.}
    \label{tab:output-comparison1}
\end{table*}

\paragraph{Quantitative Results}
In this study, three models are employed on generating scientific text sequences that can be compiled as an academic paper. The baseline model, Char-LSTM, gave promising results in terms of both quantitative and qualitative results. However, the Transformer model could not improve the baseline model results because of ineffective use of fixed sequence length. Char-LSTM also uses fixed sequence length, but it carries the residual information between sequences, whereas the Transformer does not. This disability can be tolerated in plain text generation tasks, yet the effects are severe in such tasks that the long dependencies are required like LaTeX generation. 

Although the baseline model surpassed the performance of the Transformer due to the limitations mentioned above, Transformer-XL model improved the quantitative results by allowing sequence segments to carry information one to another. Transformer-XL outperformed the rest of the models by $\sim 61\%$ on both cross-entropy error (CE) and bits-per-character (BPC). The detailed comparison of quantitative results between models is given in Table \ref{tab:cross_entropy}. Besides, the validation losses of both Char-LSTM and Transformer models converged to approximately $1.15$, while the validation loss of Transformer-XL managed to converge to $0.71$ at the end of the training. The detailed figures of the losses on the training phase can be seen in Fig. \ref{fig:training_losses}. 

\paragraph{Qualitative Results on Text Generation}
The qualitative success of the studied models' outcome is correlative to the quantitative results. The baseline model (Char-LSTM) has the ability to write syntactically correct sentences consistently, even though the training of the model is performed on character-level. For instance, each sentence generated by Char-LSTM starts with a capital letter and ends with a punctuation mark consistently. Furthermore, it has the ability to use explanatory phrases such as \textit{(blue)} as shown in Table \ref{tab:output-comparison2}. Transformer-XL also generates syntactically correct sentences, yet the sentences are better formed and semantically longer dependent than the baseline model. An example passage written by Transformer-XL is also given in Table \ref{tab:output-comparison2}. 

As a drawback most of the language models suffers from, we observed that the models might suffer from the repetition of words in the conducted experiments. Despite creating syntactically correct sequences, the repetitive outputs could occur where the semantic integrities are shallow such as the beginning sentence of the sections in the generated paper. To cope with this pitfall, the softmax temperature (as mentioned in Section \ref{sec:experimental_setup}) is experimented with different temperature values when generating text and through qualitative analysis. We found that using the values between $0.5$ and $0.75$ yielded better text generation since the values in the given interval are a good balance of confidence and diversity.

\paragraph{Qualitative Results on Syntactic Features of \LaTeX}
Char-LSTM and Transformer-XL models have the ability to use simple syntactic features of \LaTeX, such as referencing and mathematical expressions. However, the baseline model could not learn more complex features. Table~\ref{tab:output-comparison1} shows that Char-LSTM starts the table with \texttt{\textbackslash begin} statement, yet is unable to end the table with \texttt{\textbackslash end} command. This problem is caused because of the bottleneck on sequence length hyperparameter. One sentence barely fits on the character sequences when the hyperparameter is set to 100 (shown in Table \ref{tab:hyperparams}). Unlike the baseline model, Transformer-XL uses variable sequence length, thus it accomplishes to learn longer dependencies. The qualitative results also show that Transformer-XL is more successful in using complex features of \LaTeX \space documents.

\section{Conclusion and Future Work}
\label{sec:conc}
In this study, we compiled a new dataset with the \LaTeX{} source files of academic papers from arXiv.org, trained some of the recent language modeling, or in general sequence modeling methods with the new dataset, and evaluated their performance in academic paper generation. By qualitative and quantitative results, we have observed that Transformer-XL model outperformed the Char-LSTM and Transformer models.

Structured text generation, which is a super-set of academic paper generation, is an important problem for evaluating sequence models since it has more obvious dependencies than the general language modeling problems and mostly these dependencies, such as the relation between \texttt{\textbackslash begin} and \texttt{\textbackslash end}, can be tested in the generated outputs with basic context-free grammars. Developing evaluation metrics based on this idea may be an interesting future work. Moreover, we deduced that the outcomes tend to deviate from the general subject of the paper, although the sentences generated by the models have short term dependencies. Therefore, our motivation for the future work is implementing augmented architectures to sustain the coherence in the subject, similar to \cite{ijcai2018-567}, in order to generate more realistic academic papers.

\bibliographystyle{unsrt}  
\bibliography{main}

\end{document}